\documentclass[sigconf]{acmart}
\usepackage{algorithm}
\usepackage{algpseudocode}

\AtBeginDocument{%
  }

\setcopyright{acmlicensed}
\copyrightyear{2026}
\acmYear{2026}
\acmDOI{XXXXXXX.XXXXXXX}

\acmConference[ICAIF '26]{ACM International Conference on AI in Finance}{November 14--17, 2026}{Milan, Italy}
\acmISBN{978-1-4503-XXXX-X/2026/11}

\begin{document}
\raggedbottom

\title{$L_0$ Feature Selection via Monte Carlo Tree Search in High-Dimensional Financial Regression}

%% ANONYMIZED AUTHORS FOR DOUBLE-BLIND REVIEW
\author{Cameron Mouangue}
\orcid{0009-0001-1318-7965} % Ton code ORCID à 16 chiffres
\affiliation{%
  \institution{LAMSADE, Université Paris Dauphine-PSL, CNRS}
  \city{Paris}
  \country{France}
}
\email{cameron.mouangue@polytechnique.edu}

\author{Tristan Cazenave}
\orcid{0000-0003-4669-9374} 
\affiliation{%
  \institution{LAMSADE, Université Paris Dauphine-PSL, CNRS}
  \city{Paris}
  \country{France}
}
\email{tristan.cazenave@lamsade.dauphine.fr}

\begin{abstract}
Feature selection is a central problem in high-dimensional financial regression, where large numbers of engineered variables and strong correlations among predictors reduce model stability and interpretability. $L_1$ regularization (Lasso) is widely used for sparse regression, but its shrinkage can be restrictive when informative signals are spread across correlated variables. Exact sparse regression avoids this shrinkage but requires solving a combinatorial optimization problem that is computationally difficult and is typically fit to minimize in-sample rather than out-of-sample error.

This paper presents a feature selection framework based on Monte Carlo Tree Search with Generalized Nested Rollout Policy Adaptation (GNRPA). Feature selection is formulated as a sequential decision process, and candidate subsets are scored by their out-of-sample validation performance rather than their training error. The method also supports a lexicographic objective that accounts for subset size and multicollinearity during the search.

We study this approach on a contemporaneous explanatory regression task -- explaining, rather than forecasting, the weekly return of the NASDAQ-100 index from 1,072 engineered features spanning market, macroeconomic, and factor data -- since this setting isolates the value of subset selection under multicollinearity from the separate problem of predictive return forecasting. On this task, the proposed method identifies sparse feature subsets with higher out-of-sample explanatory power than an optimized Lasso baseline, reaching an out-of-sample $R^2$ of 0.6961 versus 0.6313 for Lasso. These results suggest that Monte Carlo Tree Search is a practical alternative to convex regularization for sparse \emph{explanatory} feature selection in financial regression, with potential value for factor attribution and risk decomposition, where explanatory transparency matters as much as out-of-sample fit.
\end{abstract}

%% The code below is generated by the tool at http://dl.acm.org/ccs.cfm.



\maketitle

\section{Introduction}

The growing availability of financial and macroeconomic data has increased the dimensionality of empirical financial models. Quantitative models often include hundreds or thousands of engineered variables from prices, macroeconomic indicators, technical signals, and factor data. Because many of these variables are highly correlated, feature selection is crucial for improving model interpretability, reducing estimation variance, and limiting overfitting \cite{Belsley1980}.

Lasso regression \cite{Tibshirani1996} is one of the most widely used approaches for feature selection because it produces sparse solutions via an $L_1$ penalty. However, this penalty's shrinkage can be restrictive in financial applications. Many economically meaningful signals, such as spreads, distributed lags, and momentum measures, are represented by combinations of correlated variables. In these cases, Lasso can discard relevant variables or over-shrink coefficients, potentially reducing explanatory power. Coefficient shrinkage also introduces bias into the estimated regression parameters.

Exact sparse regression, usually formulated as an $L_0$ optimization problem, avoids coefficient shrinkage by selecting a subset of predictors. Identifying the optimal subset, however, is computationally difficult because the problem is NP-hard. Existing exact and approximate solvers typically minimize penalized training error, which can lead to overfitting in noisy financial datasets where the focus is out-of-sample performance rather than in-sample fit.

This paper presents a feature selection framework based on Monte Carlo Tree Search (MCTS) \cite{Kocsis2006, Silver2016}. We adapt the Montparnasse algorithm \cite{Cazenave2026Montparnasse}—an extension of Generalized Nested Rollout Policy Adaptation (GNRPA) \cite{Cazenave2020}—to formulate sparse feature selection as a sequential decision problem. Candidate feature subsets are evaluated using validation performance during the search, so the algorithm optimizes for out-of-sample explanatory power. It also incorporates a lexicographic objective that accounts for subset size and multicollinearity, and supports both fixed-cardinality and adaptive stopping strategies.

The framework is evaluated on a high-dimensional financial dataset containing 1,072 engineered features for modeling weekly NASDAQ-100 returns. As in Section 3.1, exogenous predictors (macro, volatility, and factor variables) are observed at the same time as the target return; the task is therefore one of explanatory regression rather than out-of-sample return prediction, and reported $R^2$ values should be interpreted accordingly. Our results indicate that, on this dataset, the method identifies smaller feature subsets while improving out-of-sample explanatory performance relative to optimized convex regularization methods. The main contributions of this paper are as follows.

\begin{itemize}
    \item We formulate sparse feature selection as a sequential decision problem and apply GNRPA to approximate $L_0$ optimization using validation-guided tree search.
    \item We introduce a search framework that supports a lexicographic objective, including sparsity, multicollinearity, and endogenous stopping.
    \item We evaluate the method on a high-dimensional financial regression problem and show improved out-of-sample explanatory performance compared with optimized Lasso and related baselines.
\end{itemize}
The remainder of this paper is organized as follows. Section 2 reviews related work on sparse regression and Monte Carlo Tree Search. Section 3 presents the proposed methodology. Section 4 reports the empirical results, and Section 5 concludes.

\section{Related Work}

\subsection{Feature Selection for Financial Regression}

Feature selection has long been an important topic in financial machine learning because modern financial datasets often contain large numbers of correlated explanatory variables. Applications including empirical asset pricing, return prediction, and macro-financial modeling commonly rely on feature selection to improve model interpretability and reduce overfitting \cite{Gu2020, LopezdePrado2018, Feng2020, Kozak2020, Freyberger2020}. As the number of available market, macroeconomic, and technical indicators has increased, selecting informative predictors has become increasingly important, particularly in high-dimensional settings where multicollinearity is common \cite{Belsley1980}.

Several approaches have been proposed for this problem. Feature extraction methods such as principal component analysis reduce dimensionality but replace the original variables with latent components, creating opaque models that limit interpretability \cite{Rudin2019}. More recently, neural-network architectures such as Temporal Fusion Transformers \cite{Lim2021} and LassoNet \cite{Lemhadri2021} incorporate feature selection into the learning process itself. Although these methods improve predictive performance in many applications, they generally do not produce the globally sparse linear models that remain common in financial analysis and regulatory settings.

Sparse linear regression methods remain widely used because they retain the original explanatory variables. Lasso regression \cite{Tibshirani1996} performs feature selection using an $L_1$ penalty and has become a standard baseline in high-dimensional financial regression \cite{Chinco2019}. However, shrinkage-based methods can become unstable when predictors are highly correlated and may attenuate economically meaningful signals \cite{ZouHastie2005, Fastrich2015}. Exact subset selection, commonly formulated as an $L_0$ optimization problem, avoids coefficient shrinkage and has received renewed attention using mixed-integer optimization \cite{Bertsimas2016} and coordinate-descent algorithms \cite{Hazimeh2020}. While these methods improve the scalability of sparse regression, neither directly optimizes out-of-sample performance, as both minimize penalized training error. Furthermore, they face distinct limitations in high dimensions. Mixed-integer solvers guarantee global in-sample optimality but can quickly become computationally prohibitive. Conversely, coordinate-descent methods scale well but rely on local updates; in highly collinear financial data requiring coordinated multi-variable changes, they can become trapped in suboptimal solutions.

Recent work has also explored reinforcement learning for combinatorial optimization problems \cite{Bengio2021}. These approaches provide an alternative to deterministic optimization by learning search policies over large discrete spaces. Motivated by this direction, we formulate feature selection as a sequential decision problem and score candidate subsets by validation performance over training error.

\subsection{Monte Carlo Tree Search and GNRPA}

Monte Carlo Tree Search (MCTS) is a search algorithm for sequential decision problems that estimates the value of candidate actions using repeated simulations \cite{Coulom2006, Kocsis2006}. MCTS has been successfully applied to a variety of combinatorial optimization problems because it balances exploration and exploitation while avoiding exhaustive search \cite{Browne2012}. More recently, it has been extended beyond board games to optimization tasks with large discrete search spaces \cite{Bengio2021}.

Nested Rollout Policy Adaptation (NRPA) \cite{Rosin2011} extends rollout-based search by learning a stochastic policy that gradually increases the probability of actions appearing in high-quality solutions. Subsequent work introduced selective policies to improve search efficiency by reducing exploration of less promising trajectories \cite{Cazenave16}. Generalized Nested Rollout Policy Adaptation (GNRPA) \cite{Cazenave2020} extends this framework by incorporating temperature control and heuristic biases, making it suitable for higher-dimensional search spaces.

Given a state $s$ and the set of available actions $\mathcal{A}(s)$, the probability of selecting action $a_k$ is defined as

\begin{equation}
    \pi(a_k \mid s) =
    \frac{\exp\left(\frac{w_{s,a_k}}{\tau}+\beta_{s,a_k}\right)}
    {\sum_{a_j\in\mathcal{A}(s)}
    \exp\left(\frac{w_{s,a_j}}{\tau}+\beta_{s,a_j}\right)},
    \label{eq:gnrpa_softmax}
\end{equation}

where $W$ denotes the policy weights, $\tau$ is the temperature parameter, and $\beta$ represents heuristic domain knowledge. During policy adaptation, the weights are updated according to

\begin{equation}
    w_{s,a_j}
    \leftarrow
    w_{s,a_j}
    -
    \frac{\alpha}{\tau}
    \left(
    \pi(a_j\mid s)
    -
    \mathbb{I}_{\{a_j=a^*\}}
    \right),
    \label{eq:gnrpa_update}
\end{equation}

where $\alpha$ is the learning rate and $a^*$ denotes the action selected in the best rollout sequence.

GNRPA has shown promising results in combinatorial optimization, but its application to sparse financial regression has received limited attention. In this study, we adapt GNRPA to feature subset selection by formulating sparse regression as a sequential decision process and evaluating candidate subsets using out-of-sample validation performance.

\section{Methodology}

\begin{figure*}[t]
\centering
\includegraphics[width=0.82\textwidth]{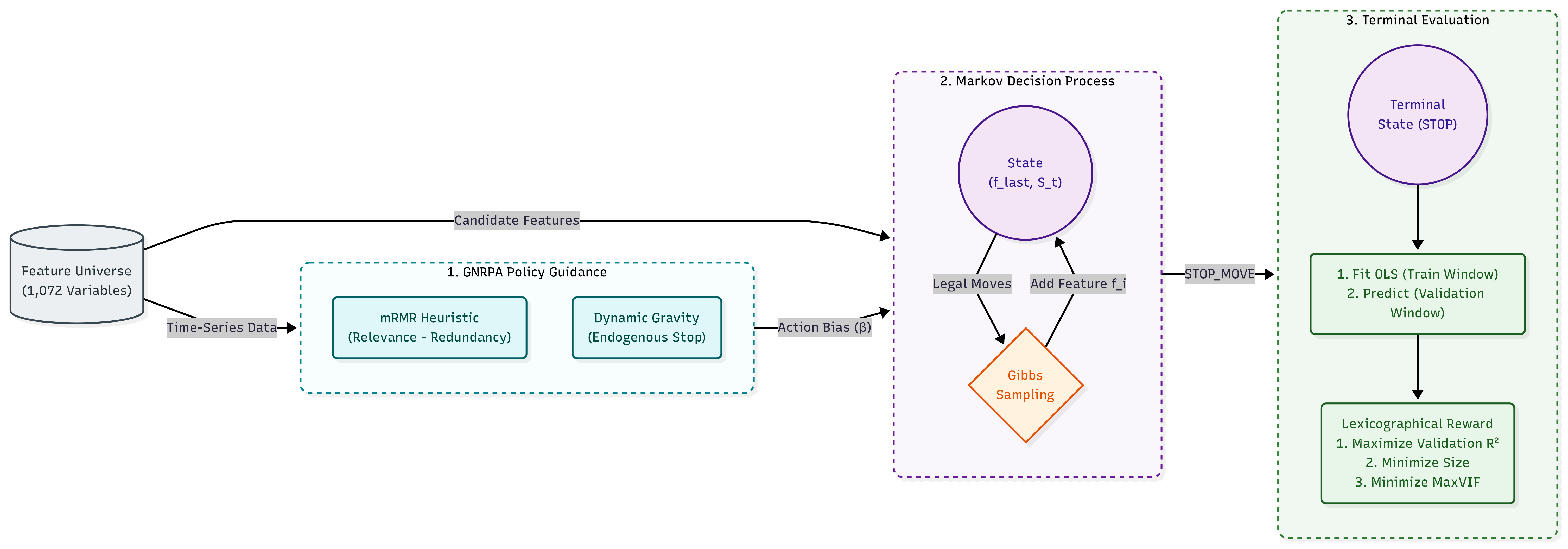}
\caption{Overview of the proposed search framework.}
\label{fig:architecture}
\end{figure*}

This section describes the feature selection framework. We first introduce the dataset and feature construction process, then formulate sparse feature selection as a Markov Decision Process (MDP). The search procedure and adaptive stopping mechanism, driven by the Montparnasse algorithm, are presented in the following subsections. Figure~\ref{fig:architecture} provides an overview of the framework.

\subsection{Dataset}

A daily financial dataset covering January 2000 through March 2026 ($T=6{,}260$ observations) was constructed from publicly available sources. The response variable is the contemporaneous weekly return of the NASDAQ-100 index, defined as the five-day rolling sum of daily returns,
\begin{equation}
Y_t=\sum_{k=0}^{4}r_{t-k},
\qquad
r_t=\frac{P_t-P_{t-1}}{P_{t-1}}.
\end{equation}
Predictor variables were obtained from three sources: (1) Market data, including price, trading volume, and volatility indices (NASDAQ-100, CBOE VIX, USD Index) via Yahoo Finance; (2) Macroeconomic data, including interest rates, yield spreads, inflation, and labor metrics via FRED; and (3) Factor data, comprising the Fama--French five factors \cite{FamaFrench2015} and momentum \cite{Carhart1997}.

The final design matrix contains $p=1{,}072$ engineered features. These include technical indicators, interaction terms, and rolling statistics computed over lookback windows $W\in\{5,21,63,126,252\}$ trading days. Technical indicators derived from the NASDAQ-100 with lookback periods shorter than five trading days were excluded to avoid information leakage. Exogenous variables, including macroeconomic indicators, volatility measures, yield spreads, and factor returns, were retained at time $t$ because they represent contemporaneously observed explanatory variables. We note that this contemporaneous alignment is deliberate rather than an oversight: $Y_t$ aggregates returns over a five-day window, while several exogenous predictors (e.g., $\Delta\text{VIX}_t$, $\Delta\text{HY Spread}_t$) are single-day changes. Consequently, the model is exploiting same-day and near-term co-movement between volatility/credit markets and equity returns---consistent with the well-documented leverage effect \cite{Black1976}---rather than a genuine lead-lag forecasting relationship. We flag this here because it bears on how the $R^2$ figures reported in Section 4 should be interpreted.

\subsection{Problem Formulation}

Sparse linear regression seeks a parameter vector
$\theta\in\mathbb{R}^{p}$ that minimizes the residual sum of squares subject to a cardinality constraint,
\begin{equation}
\min_{\theta\in\mathbb{R}^{p}}
\frac{1}{2N_{\mathrm{train}}}
\left\|
Y_{\mathrm{train}}
-
X_{\mathrm{train},S}\theta
\right\|_2^2,
\qquad
\text{s.t. }
\|\theta\|_0\le K,
\end{equation}
where $X_{\mathrm{train},S}$ denotes the design matrix restricted to the selected feature subset $S$. Candidate feature subsets are evaluated on a validation set rather than the training set.

The method supports two search modes: (1) Fixed-cardinality search, which limits subset size to a predefined $K$, and (2) Adaptive search, which uses an endogenous stopping mechanism and a lexicographic reward to determine subset size automatically.

Feature selection is formulated as an episodic Markov Decision Process (MDP) \cite{Puterman1994} defined by the tuple $(\mathcal{S},\mathcal{A},\mathcal{R})$. Figure~\ref{fig:architecture} summarizes the overall approach.

\textit{State.} Let $\mathcal{F}={f_1,\ldots,f_p}$ denote the candidate features. Although the selected subset is unordered, the search constructs it sequentially. The state is $s_t=(f_{\mathrm{last}},S_t)$, where $S_t\subset\mathcal{F}$ is the current subset and $f_{\mathrm{last}}$ is the most recently selected feature. Tracking $f_{\mathrm{last}}$ allows the $p\times p$ policy matrix to capture pairwise feature synergies, such as variables forming spread signals. The initial state is $s_0=(\emptyset,\emptyset)$.

\textit{Action.}
At each state, the agent may select any feature that is not already contained in the current subset or terminate the search,
\begin{equation}
\mathcal{A}(s_t)
=
(\mathcal{F}\setminus S_t)
\cup
\{a_{\mathrm{stop}}\}.
\end{equation}
\textit{Reward.}
Rewards are evaluated only at terminal states. For a feature subset $S$, the regression coefficients are estimated by ordinary least squares using a ridge-stabilized matrix inversion via NumPy \cite{Harris2020},
\begin{equation}
\hat{\theta}_S=
\left(
X_{\mathrm{train},S}^{\top}
X_{\mathrm{train},S}
+
\lambda I
\right)^{-1}
X_{\mathrm{train},S}^{\top}
Y_{\mathrm{train}},
\end{equation}
where $\lambda=10^{-5}$ provides numerical stability. The resulting model is evaluated on a separate validation set, and the corresponding validation performance defines the reward used during search.

\subsection{Search Heuristics}
The complete search space contains $2^{1072} \approx 10^{322}$ possible subsets. Even when constrained to a maximum capacity of $K=60$ features (the upper bound of our evaluations), the combinatorial space still exceeds $10^{99}$ valid subsets. To effectively navigate this space, we incorporate a Minimum Redundancy Maximum Relevance (mRMR) \cite{Peng2005} heuristic into the GNRPA bias term $\beta$ defined in Eq.~\eqref{eq:gnrpa_softmax}.

For a candidate feature $f_i$ and the current subset $S$, let $\rho(\cdot, \cdot)$ denote the sample Pearson correlation. The relevance $\text{Rel}(f_i)$ is defined as the absolute correlation with the target $Y$. To penalize localized correlation clusters rather than the average redundancy across the whole subset, we use a top-$k$ RMS metric. Let $\mathcal{C}_{i,S} = \{ \vert{}\rho(f_i, f_j)\vert{} : f_j \in S \}$ be the set of absolute correlations between candidate $f_i$ and the already selected features. Denoting $c_{(m)}$ as the $m$-th largest order statistic of $\mathcal{C}_{i,S}$, the redundancy is calculated over the $k$ most correlated variables:
\begin{equation}
\text{Red}_{\text{top-}k\text{-rms}}(f_i, S) = \left( \frac{1}{\min(k, \vert{}S\vert{})} \sum_{m=1}^{\min(k, \vert{}S\vert{})} c_{(m)}^2 \right)^{\frac{1}{2}}
\end{equation}
The top-$k$-RMS expression above is the general form of the redundancy term; the
search treats the redundancy metric as a categorical hyperparameter with three
settings, tuned independently per configuration (Table~\ref{tab:optuna_params}).
The \emph{mean} variant averages $c_{(m)}^2$ over all of $S$ (equivalently,
$k=|S|$ with a mean rather than root-mean-square reduction); the \emph{rms}
variant takes the root-mean-square over the full subset; and \emph{top-$k$-rms}
restricts the sum to the $k$ most correlated features as defined above. The
fixed- and tuned-cardinality configurations use the mean and rms variants
respectively, while the endogenous-stopping configuration uses top-$k$-rms
with $k=3$.
The heuristic score combines relevance and redundancy using a trade-off parameter $\gamma \in [0, 1]$, and is scaled by a magnitude parameter $\beta_{\text{mag}}$ to give the bias used in Gibbs sampling:
\begin{equation}
\beta(f_i \mid S) = \beta_{\text{mag}} \left[ \gamma \cdot \text{Rel}(f_i) - (1 - \gamma) \cdot \text{Red}_{\text{top-}k\text{-rms}}(f_i, S) \right] \label{eq:mrmr_bias}
\end{equation}
\subsection{Adaptive Search}
A limitation of standard $L_0$ implementations is the requirement to pre-specify the target cardinality $K = |S|$. We address this by allowing the search policy to determine the stopping depth automatically, so that sparsity is endogenous to the search rather than fixed in advance. Algorithm~\ref{alg:endogenous_stopping} outlines this mechanism within the search procedure.
\begin{algorithm}[b]
\small
\caption{GNRPA with Endogenous Stopping}
\label{alg:endogenous_stopping}
\begin{algorithmic}[1]
\Require Level $L$, State $s$, Policy $W$, Max Iters $I_L$, Patience $P$
\State \textbf{Global Hyperparameters:} Learning rates $\alpha_{L1}, \alpha_{\text{high}}$, Temperature $\tau$

\Function{GNRPA}{$L$, $s$, $W$}
    \If{$L == 0$} \Comment{Terminal Playout Phase}
        \State \Return \Call{Rollout}{$s, W$}
    \EndIf
    
    \State $\alpha \gets \alpha_{L1}$ \textbf{if} $L == 1$ \textbf{else} $\alpha_{\text{high}}$
    \State $best\_score \gets (-\infty, -\infty, -\infty)$
    \State $best\_seq \gets \emptyset$, \quad $no\_improve \gets 0$, \quad $n \gets 0$
    
    \While{\textbf{True}}
        \State $seq, score \gets$ \Call{GNRPA}{$L - 1$, $s, W$}
        \If{$score > best\_score$ (lexicographically)}
            \State $best\_score \gets score$
            \State $best\_seq \gets seq$
            \State $no\_improve \gets 0$
        \Else
            \State $no\_improve \gets no\_improve + 1$
        \EndIf
        
        \State $n \gets n + 1$
        
        \If{$L == 1$ \textbf{and} $no\_improve \ge P$}
            \State \textbf{break} \Comment{Early stopping}
        \EndIf
        
        \If{$L > 1$ \textbf{and} $n \ge I_L$}
            \State \textbf{break} \Comment{Fixed iterations}
        \EndIf
        
        \State Update weights $W$ toward $best\_seq$ via Eq.~\eqref{eq:gnrpa_update} using $\alpha$
    \EndWhile
    \State \Return $best\_seq, best\_score$
\EndFunction

\Statex

\Function{Rollout}{$s$, $W$}
    \While{$a_{\text{stop}}$ not selected \textbf{and} $|s| < K_{\text{max}}$}
        \State $\beta \gets$ \Call{ComputeHeuristics}{$s, \mathcal{F}$} \Comment{Eq.~\ref{eq:mrmr_bias}, \ref{eq:stop_gravity}}
        \State Sample action $a \sim \pi(a \mid s)$ using $W, \beta, \tau$ \Comment{Eq.~\ref{eq:gnrpa_softmax}}
        \State $s \gets s \cup \{a\}$ \Comment{Updates both $f_{\text{last}}$ and $S_t$}
    \EndWhile
    \State \Return $s$, \Call{EvaluateState}{$s, X_{\text{train}}, Y_{\text{train}}$} \Comment{Eq.~\ref{eq:lexicographical}}
\EndFunction
\end{algorithmic}
\end{algorithm}

To balance feature addition against termination during Gibbs sampling (Algorithm~\ref{alg:endogenous_stopping}, Line 6), the heuristic bias for $a_{\text{stop}}$ is formulated as a piecewise function that grows with the subset's proximity to the minimum required size ($K_{\text{min}}$) and the maximum capacity ($K_{\text{max}}$):
\begin{equation}
\beta(a_{\text{stop}} \mid S) = \begin{cases} -\infty & \text{if } \vert{}S\vert{} < K_{\text{min}} \\ \beta_{\text{mag}} \gamma \beta_{\text{stop}} \left( \frac{\vert{}S\vert{} - K_{\text{min}}}{\max(1, K_{\text{max}} - \vert{}S\vert{})} \right)^{\frac{3}{2}} & \text{if } \vert{}S\vert{} \ge K_{\text{min}} \end{cases} \label{eq:stop_gravity}
\end{equation}
where $\beta_{\text{stop}}$ is a hyperparameter that scales the stopping bias, and the exponent $\frac{3}{2}$ is an empirically chosen constant. This piecewise function increases the bias toward $a_{\text{stop}}$ as the subset size approaches $K_{\text{max}}$, while allowing exploration when the subset size is close to $K_{\text{min}}$.

Alongside endogenous stopping, the Montparnasse algorithm introduces a dual-learning-rate mechanism to stabilize learning in noisy environments. At the base level of the search tree, a conservative learning rate ($\alpha_{L1}$) prevents stochastic, single-rollout evaluations from prematurely skewing the policy. At higher nesting levels, where trajectories have been filtered using repeated simulations, a larger learning rate ($\alpha_{\text{high}}$) is used to reinforce proven, high-quality sequences.
Algorithm~\ref{alg:endogenous_stopping} shows the general two-rate form used for
the endogenous-stopping configuration. The fixed- and tuned-cardinality
configurations are the special case $\alpha_{L1} = \alpha_{\text{high}}$, i.e.\
a single global learning rate, consistent with the ``(global)'' entries in
Table~\ref{tab:optuna_params}. 

To avoid committing every run to a fixed number of root-level iterations, the
level-1 loop (the outermost iteration in Algorithm~\ref{alg:endogenous_stopping}
at $L=1$) is terminated early once the best lexicographic score fails to
improve for $P$ consecutive iterations. Here $P$ is the \emph{level-1 patience}
hyperparameter (Table~\ref{tab:optuna_params}). This is why the root-level
iteration count $I_1$ is reported as variable (``var.'') rather than fixed:
individual runs may stop before exhausting their maximum iteration budget once
no further improvement is observed. Level-1 patience was itself included in the
Optuna search space for the endogenous-stopping configuration.

To improve model stability, the scalar reward evaluated at the terminal state is replaced by a lexicographic tuple:
\begin{equation}
\mathcal{R}(S) = \left( \lfloor R^2_{\text{val}}(S) \rceil_3, \;-\vert{}S\vert{}, \;-\text{MaxVIF}(S) \right) \label{eq:lexicographical}
\end{equation}
where $\lfloor R^2_{\text{val}} \rceil_3$ denotes the validation $R^2$ rounded to three decimal places. Ties are broken by prioritizing smaller subsets, followed by minimizing the maximum Variance Inflation Factor (MaxVIF). MaxVIF is a standard econometric diagnostic for multicollinearity \cite{Belsley1980} and can be computed from the empirical correlation matrix $C_S \in \mathbb{R}^{\vert{}S\vert{} \times \vert{}S\vert{}}$ of the variables in subset $S$:
\begin{equation}
\text{MaxVIF}(S) = \max_{1 \le j \le \vert{}S\vert{}} \left( C_S^{-1} \right)_{jj} \label{eq:vif}
\end{equation}
By evaluating VIF as a tie-breaker inside the search tree, the algorithm penalizes highly collinear subsets that convex penalties might fail to separate.

\section{Experimental Results}

We evaluate the method to: (1) compare GNRPA-based $L_0$ selection against convex penalties to highlight shrinkage limitations; (2) assess the trade-offs between fixed, tuned, and endogenous-stopping configurations under varying search budgets; and (3) examine subset interpretability and ensemble stability. We address these objectives using an out-of-sample test on a held-out NASDAQ-100 window (January 2024--March 2026), with walk-forward cross-validation on pre-2024 data for hyperparameter tuning and validation.

\subsection{Experimental Setup and Baselines}
The proposed method was evaluated on the contemporaneous modeling task of explaining the weekly return (5-day rolling sum) of the NASDAQ-100 index. To avoid data leakage, the dataset was partitioned chronologically at December 31, 2023. The final 562 trading days (January 2024 to March 2026) were held out entirely as an out-of-sample (OOS) test set. All OOS $R^2$ values are computed using the standard coefficient of determination, where the residual variance is normalized by the variance of the test sample itself.

The search procedure used only data preceding this split date. The 1,890-day training window is not tuned for the method. It is the lookback period selected by cross-validating the Lasso baseline (described later in this section), and is reused unchanged as a fixed window for the fixed- and tuned-cardinality GNRPA configurations. Only in the final, endogenous-stopping configuration is the training window itself included in the Optuna search space (Table~\ref{tab:optuna_params}), where a shorter 1,000-day window was found to perform best jointly with the other search hyperparameters.

To compute the validation-guided reward, the search environment used a continuous block of historical data comprising the training window immediately followed by a 562-day internal validation window, positioned immediately before the 2024--2026 test period. This is deliberate rather than incidental: because market regimes shift over time, we validate the search against the period most similar in time -- and presumably most similar in regime -- to the eventual test period, mirroring the same logic used for the final train/test split itself. This trades some protection against regime change for a validation signal more representative of the conditions the final model is actually evaluated on; we return to this trade-off in Section 5. The final OLS models were then evaluated only on the post-2024 hold-out period.

Hyperparameter selection itself never accesses this hold-out period: Optuna's objective is computed by averaging validation $R^2$ across three non-overlapping walk-forward folds constructed entirely from pre-2024 data, with zero temporal overlap between folds. The post-2024 test window used for all $R^2$ figures reported in Table~\ref{tab:main_results} is therefore untouched by both the GNRPA search and the outer hyperparameter tuning loop. We note that the best-trial CV objective (Table~\ref{tab:optuna_params}, up to 0.761) exceeds the corresponding true test $R^2$ (0.643--0.668); this gap is the expected signature of the \emph{optimizer's curse} \cite{SmithWinkler2006} that arises whenever hyperparameters are chosen as the argmax over many noisy trials, and does not imply leakage, given the temporal separation of the test period.

As benchmarks, we evaluated standard penalized linear models \cite{Pedregosa2011}. Ridge regression ($L_2$ penalty) \cite{Hoerl1970} stabilized coefficient estimation but did not perform feature selection, retaining all 1,072 predictors (OOS $R^2 = 0.5250$).

For sparse feature selection, we used Lasso regression ($L_1$ penalty) \cite{Tibshirani1996}. Cross-validation selected a 1,890-day lookback window and $\lambda=0.0745$, yielding an OOS $R^2$ of 0.6313 with 60 selected features. Increasing $\lambda$ to obtain a GNRPA-comparable subset (14--16 features) reduced OOS $R^2$ to 0.39--0.43, compared with 0.6648 for GNRPA, illustrating the trade-off between sparsity and coefficient shrinkage.

To test whether combining the two penalties could address this collinearity, we evaluated an Elastic Net baseline over a grid of 180 hyperparameter combinations. Cross-validation selected an $L_1$ ratio of 0.990 (the maximum grid bound prior to pure Lasso), giving 55 features and an OOS $R^2$ of 0.6243. This suggests that the $L_2$ penalty's tendency to distribute weights across correlated variables may be suboptimal for this dataset, possibly because it restricts the large, opposing coefficients used to capture spread signals. 

As a discrete baseline for $L_0$ selection, we implemented a forward stepwise selection (greedy $L_0$) algorithm \cite{Hastie2009}. Using the same 3-fold cross-validation procedure, the algorithm stopped at $K=5$ features, giving an OOS $R^2$ of only 0.2489. 

\subsection{Explanatory Performance}

We evaluated the method across three configurations, using the Optuna framework \cite{Akiba2019} to tune the search hyperparameters. Table~\ref{tab:main_results} summarizes the comparative out-of-sample fit and statistical significance of the models.

The greedy $L_0$ baseline illustrates the limitation of myopic feature selection in a highly collinear setting, since it does not account for interactions between features chosen jointly. In the fixed-cardinality configuration, the search was constrained to select exactly 60 features per run (341 unique features across the ensemble). The mean out-of-sample $R^2$ of individual GNRPA runs (0.6428) exceeded the Lasso baseline, suggesting that MCTS is better able to identify variable combinations that myopic or shrinkage-based methods might obscure.

To estimate the underlying sparsity of the dataset, we then tuned the target cardinality with Optuna. This tuned-cardinality configuration selected 15 features per run (74 across the ensemble) and reached an OOS $R^2$ of 0.6648, suggesting that a small subset captures the primary explanatory signal.

Finally, the endogenous-stopping configuration removes the cardinality constraint; individual runs converged on an average subset size of 18.6 features (59 across the ensemble). Iteration budgets ($I_1 \times I_2 \times I_3$) were scaled across configurations based on empirical convergence observed during preliminary tuning: while the fixed configuration plateaued quickly, the richer learning dynamics of the endogenous mechanism benefited from deeper exploration. Endogenous stopping combined with an additional exploration budget and the lexicographic reward reduced the variance between runs relative to the tuned-cardinality configuration. The tuned-cardinality configuration is faster and reaches a similar ensemble $R^2$ (mean $0.6648$, $p < 0.01$ vs.\ Lasso via one-sample $t$-test), though its worst run drops to $0.519$. Endogenous stopping reaches a mean $R^2$ of $0.6682$ ($p < 0.001$ vs.\ Lasso), and even its worst run ($0.650$, Table~\ref{tab:ensemble_convergence_detailed}) beats Lasso.

Most of the additional search time for the endogenous-stopping configuration (1671.6 seconds, versus <1s for Lasso) appears to come from the level-1 patience hyperparameter, which controls exploration at the root of the search, rather than from the stopping mechanism itself. The results therefore illustrate a trade-off between computational cost, subset size, and variation across runs: a single endogenous-stopping run gives a sparse model (18.6 features on average) that already outperforms Lasso, while combining runs into an ensemble by soft voting gives the highest out-of-sample $R^2$ (0.6961) at the cost of a larger combined feature set (59 variables).
{
\setlength{\textfloatsep}{1pt}
\setlength{\intextsep}{1pt}
\begin{table}[t]
    \centering
    \footnotesize
    \setlength{\tabcolsep}{4pt}
    \caption{Out-of-sample explanatory performance, execution time, and sparsity over the 562-day test window.}
    \label{tab:main_results}
    \resizebox{\columnwidth}{!}{%
    \begin{tabular}{@{}lcccc@{}}
        \toprule
        \textbf{Model} & \textbf{Feat.} & \textbf{Time (s)} & \textbf{Mean Indiv. $R^2$} & \textbf{Ens. $R^2$} \\
        \midrule
        \multicolumn{5}{l}{\textit{Baselines}} \\
        Ridge ($L_2$) & 1,072 & $< 1.0$ & 0.5250 & -- \\
        Lasso ($L_1$) & 60 & $< 1.0$ & 0.6313 & -- \\
        Elastic Net & 55 & $< 1.0$ & 0.6243 & -- \\
        Greedy ($L_0$) & 5 & 17.1 & 0.2489 & -- \\
        \midrule
        \multicolumn{5}{l}{\textit{Proposed framework}} \\
        Fixed cardinality & 60 & 52.9 & 0.6428 & 0.6855 \\
        Tuned cardinality & 15 & 106.9 & 0.6648 & 0.6908 \\
        \textbf{Endogenous stopping} & \textbf{18.6} & \textbf{1671.6} & \textbf{0.6682} & \textbf{0.6961} \\
        \bottomrule
    \end{tabular}%
    } 
\end{table}
}
\subsection{Implementation Details and Hyperparameter Optimization}

Hyperparameter tuning was conducted with the Optuna framework \cite{Akiba2019}; optimal values for all three configurations are reported in Table~\ref{tab:optuna_params}. Optuna budgets were 15, 50, and 20 trials for the fixed-, tuned-, and endogenous-stopping configurations respectively, corresponding to total hyperparameter-search wall-clock times of 0.41, 4.08, and 11.27 hours on a 16-core workstation (in addition to the ensemble-building time in Table~\ref{tab:main_results}). The optimal hyperparameters differed across configurations: the fixed-cardinality configuration required a higher temperature ($\tau=1.520$) for broader exploration, while the tuned-cardinality configuration favored more exploitation ($\tau=0.868$). For the endogenous-stopping configuration, the training window itself was included in the search space, unlike the other two configurations, where it is fixed to the window selected by the Lasso baseline (Section 4.1). This configuration converged to a shorter 1,000-day trailing period, with a two-level learning rate: small at the root ($\alpha_{L1}=0.177$) and larger at deeper levels ($\alpha_{\text{high}}=2.212$).

To improve the reliability of the results, we used an ensemble of independent runs. Runs were executed in batches of six and combined by averaging (soft voting). The ensemble was extended until the aggregate validation $R^2$ stabilized (change below 0.01 across two consecutive batches). Because Lasso and Elastic Net are deterministic given a fixed regularization strength, they admit no equivalent notion of an ensemble. Consequently, Table~\ref{tab:main_results} compares these single models against both the expected performance of a single stochastic-search run and the combined soft-voted ensemble. Table~\ref{tab:ensemble_convergence_detailed} reports the full ensemble statistics and the number of unique features selected globally.

\subsection{Interpretability and Ensemble Stability}

Interpretability is an important property for financial models. As the Elastic Net results suggest, convex penalties can struggle to preserve interactions between variables because they penalize the size of the coefficients. Because $L_0$ selection limits only the number of features and applies no shrinkage, OLS can assign large, opposing weights to correlated variables to construct spread signals.

\begin{figure*}[t!]
    \centering
    \noindent
    \footnotesize % Forces the EXACT same font size across all three tables
    \setlength{\tabcolsep}{4pt} % Tightens horizontal whitespace so wide tables fit naturally
    
    % --- TOP ROW: TWO COLUMNS ---
    % Left Column (66% width): Contains Table 2 and Table 3 stacked
    \begin{minipage}[t]{0.64\textwidth}
        \centering
        
        % TABLE 2
        \captionof{table}{Optimal hyperparameters per configuration, selected via Optuna on pre-2024 data only.}
        \label{tab:optuna_params}
        \vspace{-0.1cm}
        \renewcommand{\arraystretch}{1.05}
        \begin{tabular}{lccc}
            \toprule
            \textbf{Hyperparameter} & \textbf{Fixed card.} & \textbf{Tuned card.} & \textbf{Endogenous stopping} \\
            \midrule
            Training window (days) & 1,890 (from Lasso CV) & 1,890 (from Lasso CV) & 1,000 (GNRPA-tuned) \\
            Target features $K$ & 60 (fixed) & 15 & endogenous (mean 18.6) \\
            Redundancy metric & mean & rms & top-3-rms \\
            Trade-off $\gamma$ & 0.666 & 0.772 & 0.811 \\
            Magnitude $\beta_{\text{mag}}$ & 24.82 & 14.97 & 13.72 \\
            Stop gravity $\beta_{\text{stop}}$ & -- & -- & 0.909 \\
            Temperature $\tau$ & 1.520 & 0.868 & 1.515 \\
            Learning rates ($\alpha_{L1}$, $\alpha_{\text{high}}$) & 1.514 (global) & 2.003 (global) & 0.177, 2.212 \\
            Tree iters.\ ($I_1{\times}I_2{\times}I_3$) & $84{\times}3{\times}6$ & $18{\times}28{\times}18$ & var.${\times}30{\times}15$ \\
            Level-1 patience & -- & -- & 63 \\
            \midrule
            \textbf{Best CV validation $R^2$} & \textbf{0.7089} & \textbf{0.7491} & \textbf{0.7607} \\
            \bottomrule
        \end{tabular}

        \vspace{0.4cm} % Space between the stacked tables

        % TABLE 3
        \captionof{table}{Ensemble diagnostics (soft-voted average of independent runs).}
        \label{tab:ensemble_convergence_detailed}
        \vspace{-0.1cm}
        \renewcommand{\arraystretch}{1.15}
        \begin{tabular}{lccccc}
            \toprule
            \textbf{Model} & \textbf{Runs} & \textbf{Ens.\ build time (s)} & \textbf{Val $R^2$} & \textbf{Indiv.\ OOS $R^2$ ($\mu\pm\sigma$, range)} & \textbf{Global feat.} \\
            \midrule
            Fixed cardinality & 18 & 166 & 0.702 & $0.643\pm0.030$ $[0.575,0.686]$ & 341 \\
            Tuned cardinality & 18 & 343 & 0.723 & $0.665\pm0.039$ $[0.519,0.698]$ & 74 \\
            \textbf{Endogenous stopping} & \textbf{12} & \textbf{2127} & 0.755 & $0.668\pm0.009$ $[0.650,0.682]$ & \textbf{59} \\
            \bottomrule
        \end{tabular}
        
    \end{minipage}\hfill
    % Right Column (28% width): Contains Table 4
    \begin{minipage}[t]{0.33\textwidth}
        \centering
        
        % TABLE 4
        \captionof{table}{Composition of the 15-feature subset analyzed above}
        \label{tab:golden_portfolio}
        \vspace{-0.1cm}
        \begin{tabular}{lr}
            \toprule
            \textbf{Selected Feature} & \textbf{Coefficient ($\beta$)} \\
            \midrule
            \multicolumn{2}{l}{\textit{Volatility Metrics}} \\
            $\text{VIX}_{t-5}$ & +0.0513 \\
            $\text{VIX}_{t-1}^{(5\text{d mean})}$ & -0.0621 \\
            $\text{VIX}_{t}^2$ & +0.0109 \\
            $\Delta\text{VIX}_{t}$ & -0.0095 \\
            $\Delta\text{VIX}_{t-1}$ & -0.0043 \\
            $\Delta\text{VIX}_{t-2}$ & -0.0023 \\
            \midrule
            \multicolumn{2}{l}{\textit{Credit Stress}} \\
            $\Delta\text{HY Spread}_{t}$ & -0.0044 \\
            $\Delta\text{HY Spread}_{t-1}$ & -0.0052 \\
            $\Delta\text{HY Spread}_{t-2}$ & -0.0046 \\
            $\Delta\text{HY Spread}_{t-3}$ & -0.0041 \\
            \midrule
            \multicolumn{2}{l}{\textit{Factor Exposures}} \\
            $\text{CMA}_{t}^{(5\text{d mean})}$ & -0.0056 \\
            $\text{CMA}_{t-1}^{(5\text{d mean})}$ & +0.0005 \\
            $\text{HML}_{t}^{(5\text{d mean})}$ & -0.0080 \\
            $\text{HML}_{t-2}^{(5\text{d std})}$ & +0.0024 \\
            $\text{SMB}_{t}^{(21\text{d mean})}$ & +0.0014 \\
            \bottomrule
        \end{tabular}
        
    \end{minipage}

    \vspace{0.6cm} % Space between the top tables and the bottom figure

    \includegraphics[width=0.82\textwidth]{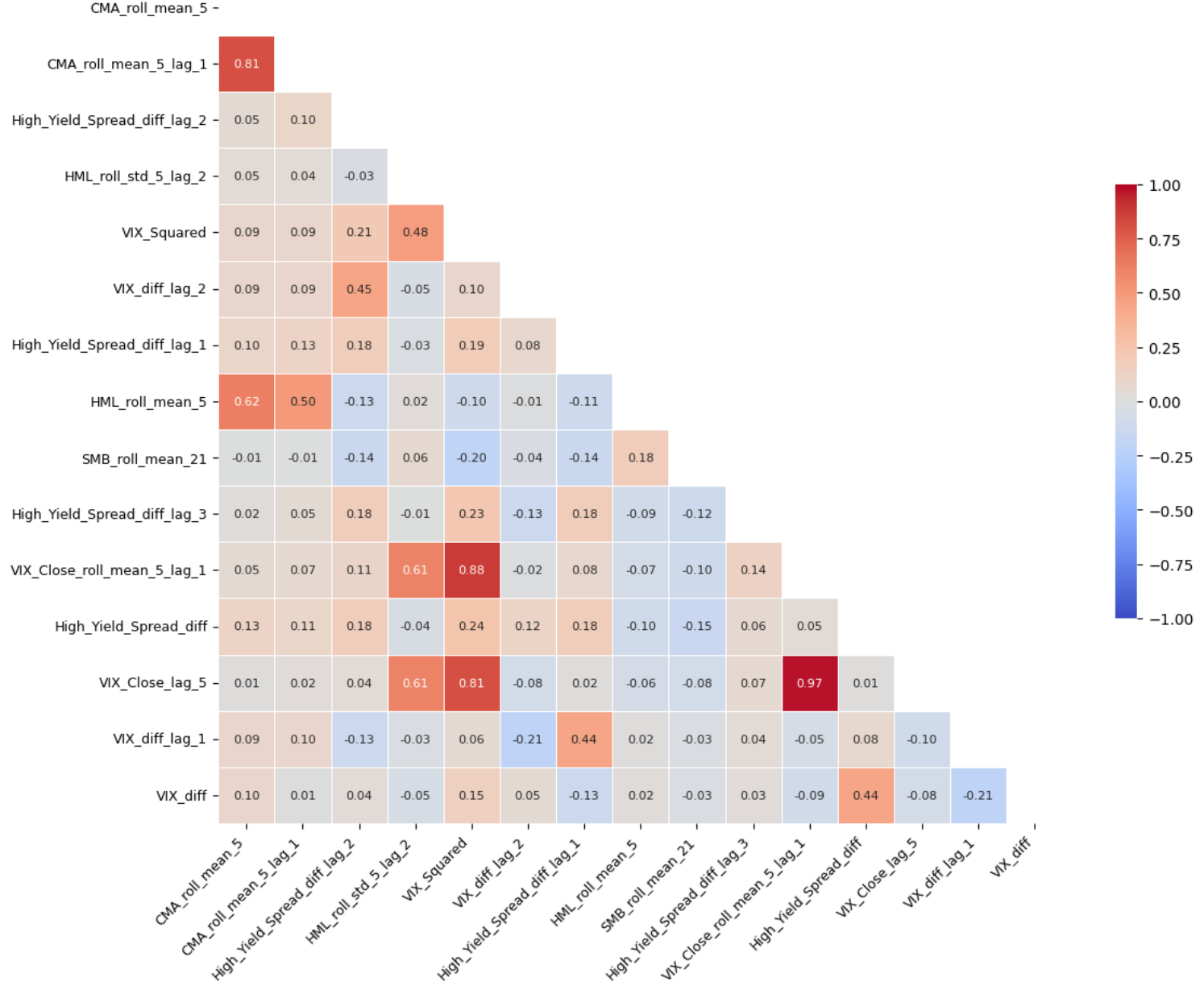}
    \caption{Correlation matrix of the best performing individual 15-feature GNRPA subset.}
    \label{fig:corr_gnrpa}
\end{figure*}

To check this intuition, we inspected the best-performing individual run from the tuned-cardinality ensemble (15 features, OOS $R^2=0.6980$). We highlight this specific run qualitatively to illustrate the underlying mechanism, not as our headline result; treating a post hoc ``best draw'' as the primary metric conflates expected performance with selection bias. Consequently, the ensemble-averaged $R^2$ in Table~\ref{tab:main_results} remains our primary evaluation metric throughout this paper, reflecting the reliable performance an independent user can expect. Figure~\ref{fig:corr_gnrpa} and Table~\ref{tab:golden_portfolio} show that this subset forms an interpretable distributed-lag model spanning several economic mechanisms.

The selected variables suggest economic mechanisms that continuous penalties may attenuate.
(1) \textit{Volatility spread construction:} The subset pairs the rolling mean ($\text{VIX}_{t-1}^{(5\text{d mean})}$, $\beta=-0.0621$) with a longer lag ($\text{VIX}_{t-5}$, $\beta=+0.0513$) consistent with capturing short-term volatility persistence and mean reversion. Continuous penalties shrink these opposing coefficients, which can attenuate the pattern, whereas unshrunk OLS preserves it. 
(2) \textit{Credit spreads:} The daily change in the High Yield spread ($\Delta\text{HY Spread}_{t}$) is selected alongside its first three lags. All four coefficients are negative and similarly sized ($\beta \in [-0.0052,-0.0041]$), capturing a cumulative drag where credit spread changes affect returns over consecutive days. 
(3) \textit{Factor exposures:} The Conservative-Minus-Aggressive factor ($\text{CMA}_{t}^{(5\text{d mean})}$, $\beta=-0.0056$) is included alongside its one-day lag ($\beta=+0.0005$). The search retains both, with the lag appearing to act as a temporal smoothing term.

To reduce the variance of the stochastic search, we averaged the predictions of the independent OLS regressions (soft voting). Unlike a tree-based ensemble, the average of $M$ linear regressions is itself a single linear equation:
\begin{equation}
    \hat{y}_{\text{ens}} = \frac{1}{M} \sum_{m=1}^{M} X \hat{\theta}^{(m)} = X \left( \frac{1}{M} \sum_{m=1}^{M} \hat{\theta}^{(m)} \right) = X \theta^*
\end{equation}
which preserves linear performance attribution. The endogenous-stopping ensemble used 59 unique variables in total out of the 1,072 candidate features, fewer than the 74 unique features used by the tuned-cardinality ensemble, even though its individual runs selected slightly larger subsets. This suggests that endogenous stopping leads independent runs to converge on a more consistent set of features, yielding an out-of-sample $R^2$ of 0.6961. Future work could formalize this consistency using stability selection metrics, such as selection frequency and inter-run Jaccard similarity \cite{Meinshausen2010}.

\section{Discussion and Conclusion}

This paper presents a feature selection framework, based on reinforcement learning, for high-dimensional financial regression. By formulating validation-guided $L_0$ subset selection as a sequential Markov Decision Process, the approach avoids the shrinkage bias of convex regularization methods such as Lasso, while retaining the interpretability of linear models \cite{Rudin2019, LopezdePrado2018}.

The contributions of this study are threefold. First, it introduces an adaptive search framework that optimizes discrete subsets for out-of-sample performance, avoiding convex shrinkage. Second, it demonstrates that Monte Carlo Tree Search readily accommodates domain-specific feature selection objectives, using the maximum Variance Inflation Factor to favor subsets with lower internal multicollinearity when explanatory performance is similar. Third, it shows that the method finds sparse, interpretable feature subsets that outperform the convex baselines out of sample on this dataset, and that a linear ensemble of these subsets reaches an out-of-sample $R^2 = 0.6961$ while preserving additive performance attribution.

The improved out-of-sample fit comes at a higher computational cost and tuning burden than convex baselines, though reported Python runtimes on a single workstation are conservative and could be reduced via compiled languages, parallelized ensembles, and ablating the root-level patience parameter $P$.

Finally, while our evaluation uses a single index and out-of-sample window (January 2024--March 2026), the consistent validation and test gains over convex baselines support the method's efficacy. Extending this benchmark across rolling test windows, additional asset classes, and non-convex penalties (e.g., SCAD, MCP) remains an important direction for future work.

A natural extension of this work is applying MCTS to other combinatorial problems in finance, such as cardinality-constrained index tracking \cite{Fastrich2015}. Unlike traditional metaheuristics such as genetic algorithms or simulated annealing \cite{Chang2000}, which mutate complete candidate solutions, GNRPA builds a solution incrementally and adapts its policy during the search. Comparing these approaches under long-only and fully-invested portfolio constraints would help clarify the practical value of reinforcement learning for this class of problem in quantitative finance.

%% Bibliography
\bibliographystyle{ACM-Reference-Format}
\bibliography{sample-base} 

\end{document}